\documentclass[conference]{IEEEtran}
\IEEEoverridecommandlockouts

\usepackage{cite}
\usepackage{orcidlink}
\usepackage{float}
\usepackage{booktabs}
\usepackage{subfigure}
\usepackage{subcaption}
\usepackage{lettrine}
\usepackage{amsmath,amssymb,amsfonts}
\usepackage{algorithmic}
\usepackage{graphicx}
\usepackage{textcomp}
\usepackage{xcolor}

\def\BibTeX{{\rm B\kern-.05em{\sc i\kern-.025em b}\kern-.08em
		T\kern-.1667em\lower.7ex\hbox{E}\kern-.125emX}}

\begin{document}
	
	\title{Wideband Power Amplifier Behavioral Modeling Using an Amplitude Conditioned LSTM}
	
	\author{
		\IEEEauthorblockN{Abdelrahman Abdelsalam\orcidlink{0009-0003-2719-8774}}
		\IEEEauthorblockA{\textit{School of Electronic Science and Engineering} \\
			\textit{University of Electronic Science and Technology of China} \\
			Chengdu, China \\
			abdelsalmabdelrahman0@gmail.com}
		\and
		\IEEEauthorblockN{You Fei}
		\IEEEauthorblockA{\textit{School of Electronic Science and Engineering} \\
			\textit{University of Electronic Science and Technology of China} \\
			Chengdu, China \\
			feiyou@uestc.edu.cn}
	}
	
	\maketitle
	
	\begin{abstract}
		Wideband power amplifiers exhibit complex nonlinear and memory effects that challenge traditional behavioral modeling approaches. 
		This paper proposes a novel amplitude-conditioned long short-term memory (AC-LSTM) network that introduces explicit amplitude-dependent gating to enhance the modeling of wideband PA dynamics. 
		The architecture incorporates a Feature-wise Linear Modulation (FiLM) layer that conditions the LSTM's forget gate on the instantaneous input amplitude, providing a physics-aware inductive bias for capturing amplitude-dependent memory effects. 
		Experimental validation using a 100 MHz 5G NR signal and a GaN PA demonstrates that the proposed AC-LSTM achieves a normalized mean square error (NMSE) of -41.25 dB, representing a 1.15 dB improvement over standard LSTM and 7.45 dB improvement over augmented real-valued time-delay neural network (ARVTDNN) baselines. 
		The model also closely matches the measured PA's spectral characteristics with an adjacent channel power ratio (ACPR) of -28.58 dB. 
		These results shows the effectiveness of amplitude conditioning for improving both time-domain accuracy and spectral fidelity in wide-band PA behavioral modeling.
	\end{abstract}
	
	
	
	\section{Introduction}
	
	\lettrine{P}{ower} amplifiers (PAs) are critical components at the front end of most radio frequency (RF) systems, including wireless communications and radar. Their performance directly dictates the range, efficiency, and signal integrity of these systems. However, PAs inherently exhibit nonlinear behavior and memory effects, where the current output depends not only on the instantaneous input but also on its past values. This behavior is profoundly influenced by signal characteristics such as bandwidth, peak-to-average power ratio (PAPR), modulation scheme, and load conditions. Accurately modeling these complex dynamics is essential for designing effective linearization techniques to maintain spectral efficiency and comply with regulatory standards.
	
	To address these nonlinearities, a variety of behavioral models have been developed. Classical approaches include the Memory Polynomial (MP), the Envelope Memory Polynomial (EMP), the Generalized Memory Polynomial (GMP), and the Volterra series. While these models have been widely adopted, they face significant limitations in wideband scenarios. Polynomial-based models often require a high number of coefficients to capture strong memory effects and deep nonlinearities, leading to increased complexity, numerical instability, and poor generalization beyond their training data. The trade-off between model accuracy and implementable complexity becomes a critical bottleneck for next-generation wideband systems, such as those in 5G and beyond.
	
	Recent advances have demonstrated the superior potential of data-driven neural network (NN) models to overcome these limitations. Architectures such as Augmented Real-Valued Time-Delay Neural Networks (ARVTDNNs), Long Short-Term Memory (LSTM) networks, Bidirectional LSTMs (BiLSTMs), and Gated Recurrent Units (GRUs) have shown promise in accurately modeling wideband PAs. These models learn the nonlinear mapping from measured input and output data, offering greater flexibility and accuracy than traditional polynomial-based methods. Subsequently, the trained model can be integrated into system simulations to fine-tune parameters or serve as a plant model for designing digital predistortion (DPD) linearizers. DPD is a crucial technique to counteract PA nonlinearities, thereby improving overall transmitter performance by pre-distorting the input signal to compensate for the amplifier's distortion.
	Concurrently, open-source benchmarking frameworks such as \textbf{OpenDPD} have been introduced to standardize dataset acquisition and enable fair comparison of DPD models across different architectures \cite{b8}.\\
	Despite their success, conventional neural network approaches, including standard LSTMs, often lack an explicit mechanism to condition their internal processing on the primary driver of PA nonlinearity: the instantaneous amplitude of the input signal. This often results in an inefficient and less accurate characterization of modern PAs amplitude-dependent memory dynamics.
	
	The main contributions of this work are threefold:
	\begin{itemize}
		\item We propose a novel \emph{amplitude-conditioned LSTM cell} for behavioral modeling of wideband power amplifiers. The core innovation is the modulation of the LSTM's forget gate using a feature-wise linear modulator (FiLM) conditioned directly on the input signal amplitude.
		\item This architecture introduces a physics-aware inductive bias, enabling the network to dynamically adapt its memory retention based on the PA's immediate operating point. This leads to a more precise and parameter-efficient modeling of complex nonlinear dynamics.
		\item We provide a comprehensive evaluation, demonstrating that the proposed model achieves superior accuracy in modeling measured PA data compared to benchmark models, including memory polynomials and standard recurrent neural networks, while maintaining competitive computational complexity.
	\end{itemize}
	
	\section{Background and Related Work}
	
	The accurate behavioral modeling of power amplifiers (PAs) is critical for designing efficient linearization systems in modern wideband communication standards like 5G and beyond. As signal bandwidths increase, PAs exhibit stronger memory-dependent nonlinearities, rendering traditional polynomial-based models (e.g., Volterra series, memory polynomials) less effective due to their limited capacity and high parameter sensitivity. This challenge has catalyzed significant research into data-driven neural network (NN) models, which offer superior universal approximation capabilities for complex nonlinear systems with memory.
	
	\subsection{Classical and Foundational Neural Network Models}
	Early investigations established the viability of NNs for PA modeling. A foundational review by Yan et al. surveyed neural network techniques specifically for capturing PA memory effects, setting a baseline for architectural comparisons \cite{b1}. Building on static Multi-Layer Perceptron (MLP) concepts, Time-Delay Neural Networks (TDNNs) incorporated memory via explicit time-delayed input taps. Yin et al. provided valuable analysis into the iteration process of such real-valued TDNNs with different activation functions, offering insights into their convergence behavior during PA behavioral modeling \cite{b3}. These works demonstrated the potential of NNs but also highlighted limitations in modeling long-term temporal dependencies inherent in wideband PAs.
	
	\subsection{Advanced Architectures for Memory Modeling}
	To explicitly capture dynamic memory effects, research shifted towards recurrent neural network (RNN) architectures. Chen et al. pioneered the application of Long Short-Term Memory (LSTM) networks for GaN PA behavioral modeling, leveraging their internal gating mechanism to overcome the vanishing gradient problem of simple RNNs and effectively model long-term dependencies \cite{b4}. This direction was advanced by Sun et al., who employed Bidirectional LSTMs (BiLSTMs) for wideband RF PAs in 5G systems. Their work explicitly linked the PA's memory effects to the BiLSTM's ability to process temporal sequences in both forward and backward directions, achieving excellent modeling accuracy \cite{b6}. These studies cemented LSTM-based networks as a dominant and effective approach for memory-heavy PA modeling.
	
	\subsection{Methodological Innovations and Efficiency}
	Concurrent with architectural advances, novel training methodologies and efficiency concerns have become prominent. Tarver et al. proposed an innovative direct learning framework for digital predistortion (DPD), where backpropagation is performed through a neural network model of the PA. This approach bypasses the traditional indirect learning architecture and co-optimizes the model and linearizer, showing improved performance \cite{b2}. As model complexity grew, the pursuit of lightweight, implementable solutions gained traction. Wu et al. directly addressed this by proposing a lightweight deep neural network based on Bidirectional Gated Recurrent Units (BGRUs), aiming to balance accuracy with a reduced parameter count for practical deployment \cite{b7}. Similarly, Khawam et al. explored a hybrid approach, combining memoryless polar domain functions with a Deep Neural Network (DNN) to model GaN Doherty PAs, effectively simplifying the neural network's required task \cite{b5}.
	
	\subsection{Research Gap and Contribution}
	The literature reveals a clear evolution from foundational MLPs to sophisticated, memory-aware LSTMs, with a growing emphasis on training efficiency and model lightweighting. However, a key gap remains: existing LSTM gates operate on learned internal states but are not dynamically \emph{conditioned} on the instantaneous input signal characteristics ,the primary driver of PA nonlinearity. Standard LSTMs must learn to infer the operating point indirectly, which can limit efficiency and precision.
	
	This paper directly addresses this gap by introducing a \emph{novel amplitude-conditioned LSTM cell} for PA behavioral modeling. Our core innovation is the modulation of the LSTM's forget gate output via a Feature-Wise Linear Modulation (FiLM) layer conditioned on the input signal's amplitude. This provides the network with an explicit, physics-aware inductive bias, allowing it to dynamically adjust memory retention based on the immediate PA operating point. By integrating this amplitude-conditioned mechanism, the proposed model promises more precise and efficient modeling of amplitude-dependent memory effects, advancing beyond the capabilities of the standard and bidirectional LSTMs reviewed in \cite{b4, b6} and contributing to the field's pursuit of high-accuracy, efficient models as seen in \cite{b5, b7}.
	
	\section{Proposed Amplitude-Conditioned LSTM Model}

	\begin{figure}[t]
		\centering
		\includegraphics[width=\columnwidth]{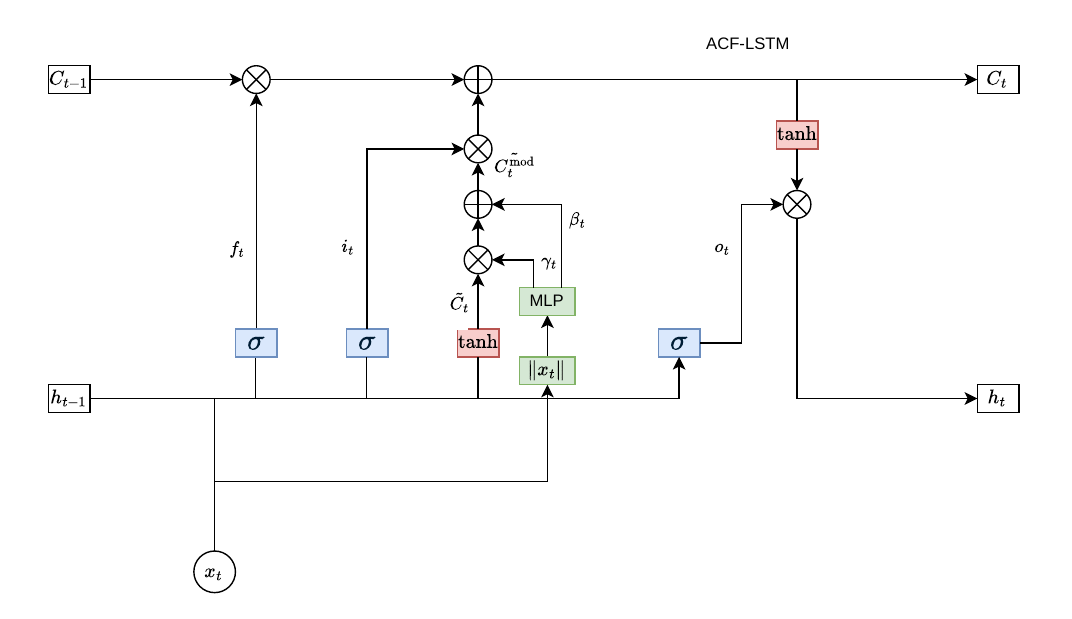} 
		\caption{The Proposed Amplitude Conditioned LSTM cell}
		\label{fig:acflstm}
	\end{figure}
	
	\begin{figure*}[t]
		\centering
		\includegraphics[width=\textwidth]{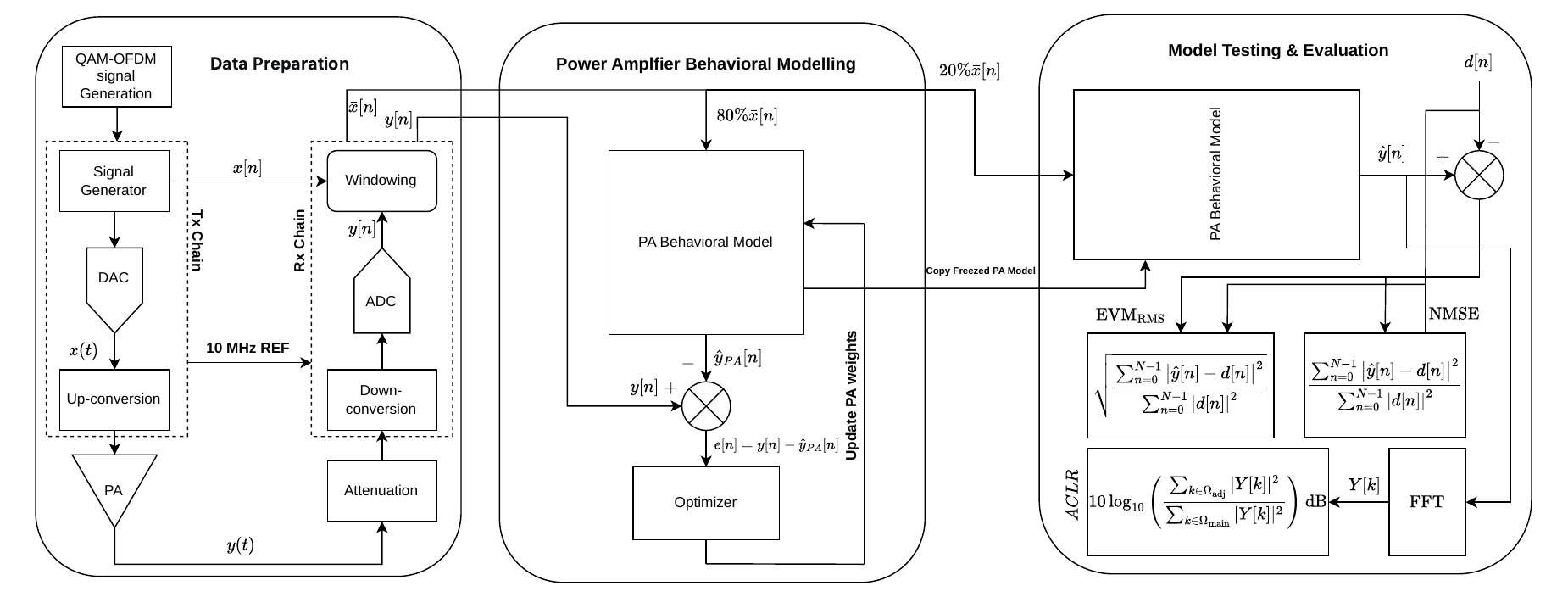} 
		\caption{Three-stage workflow for PA behavioral modeling: (a) Data Acquisition \& Pre-processing, (b) Model Training, and (c) Testing \& Evaluation.}
		\label{fig:system_workflow}
	\end{figure*}

	\begin{figure}[t]
		\centering
		\includegraphics[width=\columnwidth]{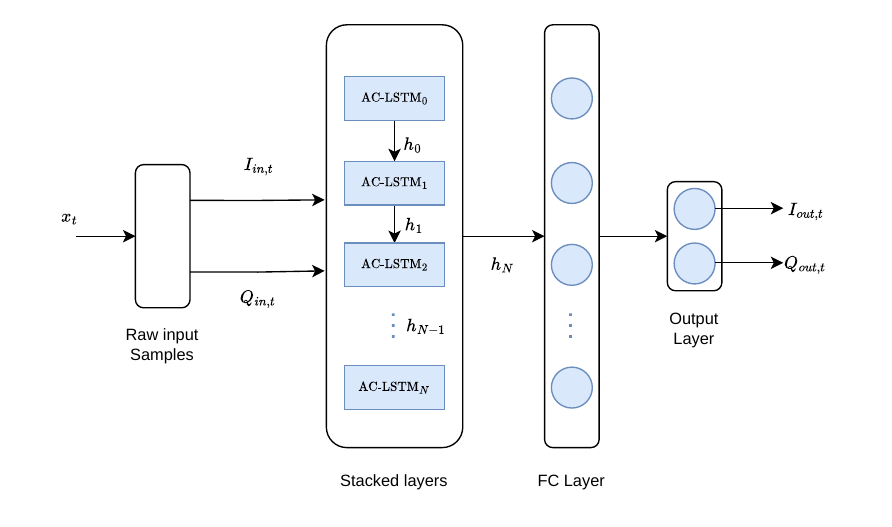} 
		\caption{General Network Architecture}
		\label{fig:network_architecture}
	\end{figure}
	
	\subsection{Model Overview}
	
	The proposed behavioral model is based on a deep recurrent neural network composed of multiple stacked amplitude-conditioned LSTM (AC-LSTM) layers, followed by a fully connected (FC) projection and a linear output layer, as illustrated in Fig.~\ref{fig:network_architecture}. The model directly processes raw complex baseband input samples and predicts the corresponding output I/Q components of the power amplifier.
	
	At each discrete time index $t$, the complex-valued input signal $x_t$ is represented in terms of its in-phase and quadrature components as
	\begin{equation}
		x_t = I_{\text{in},t} + j Q_{\text{in},t}.
	\end{equation}
	The input sequence is fed into a stack of $N$ AC-LSTM layers, where each layer captures nonlinear memory effects over time. The hidden state of the final AC-LSTM layer, denoted as $h_N$, is subsequently projected through a fully connected layer whose dimensionality matches the hidden state size. Finally, a linear output layer maps the FC output to a two-dimensional vector corresponding to the predicted in-phase and quadrature components of the PA output signal,
	\begin{equation}
		\hat{y}_t = 
		\begin{bmatrix}
			I_{\text{out},t} \\
			Q_{\text{out},t}
		\end{bmatrix}.
	\end{equation}
	
	This end-to-end architecture enables direct modeling of wideband PA behavior without explicit basis expansion or predefined memory structures.
	
	\subsection{Amplitude-Conditioned LSTM Cell}
	
	The core building block of the proposed network is the amplitude-conditioned LSTM (AC-LSTM) cell shown in Fig.~\ref{fig:acflstm}. Unlike a conventional LSTM, the proposed cell explicitly incorporates the instantaneous amplitude of the input signal to modulate its internal memory update mechanism.
	
	Specifically, the envelope of the complex input signal is computed as
	\begin{equation}
		a_t = \lVert x_t \rVert.
	\end{equation}
	This amplitude information is passed through a lightweight multi-layer perceptron (MLP) to generate feature-wise scaling and bias parameters that adaptively control the candidate cell state. By conditioning the memory update on the signal amplitude, the AC-LSTM is able to capture amplitude-dependent nonlinearities and long-term memory effects commonly observed in wideband RF power amplifiers, such as thermal and electrical memory.
	
	This conditioning mechanism allows the recurrent dynamics to vary with the instantaneous operating point of the PA, providing a physically meaningful inductive bias absent in standard recurrent architectures.
	
	\subsection{Mathematical Formulation}
	
	Let $h_{t-1}$ and $C_{t-1}$ denote the hidden and cell states of the AC-LSTM at time $t-1$, respectively. The gate activations follow the standard LSTM formulation:
	\begin{align}
		f_t &= \sigma(W_f x_t + U_f h_{t-1} + b_f), \\
		i_t &= \sigma(W_i x_t + U_i h_{t-1} + b_i), \\
		o_t &= \sigma(W_o x_t + U_o h_{t-1} + b_o),
	\end{align}
	where $\sigma(\cdot)$ denotes the sigmoid activation function.
	
	The candidate cell state is computed as
	\begin{equation}
		\tilde{C}_t = \tanh(W_c x_t + U_c h_{t-1} + b_c).
	\end{equation}
	
	To incorporate amplitude-dependent modulation, the instantaneous input amplitude $a_t$ is processed by an MLP to generate feature-wise scaling and bias vectors:
	\begin{equation}
		[\gamma_t, \beta_t] = \mathrm{MLP}(a_t).
	\end{equation}
	The candidate cell state is then modulated according to
	\begin{equation}
		\tilde{C}_t^{\mathrm{mod}} = \gamma_t \odot \tilde{C}_t + \beta_t,
	\end{equation}
	where $\odot$ denotes element-wise multiplication.
	
	The cell and hidden state updates are given by
	\begin{align}
		C_t &= f_t \odot C_{t-1} + i_t \odot \tilde{C}_t^{\mathrm{mod}}, \\
		h_t &= o_t \odot \tanh(C_t).
	\end{align}
	
	Compared to a conventional LSTM, the proposed AC-LSTM introduces amplitude-dependent modulation exclusively in the candidate memory update, allowing the model to adapt its temporal dynamics to the instantaneous signal envelope while preserving the stability and training properties of standard LSTM architectures.
	
	\section{Experimental Setup}
     
     \begin{figure}[t]
     	\centering
     	\includegraphics[width=0.85\columnwidth]{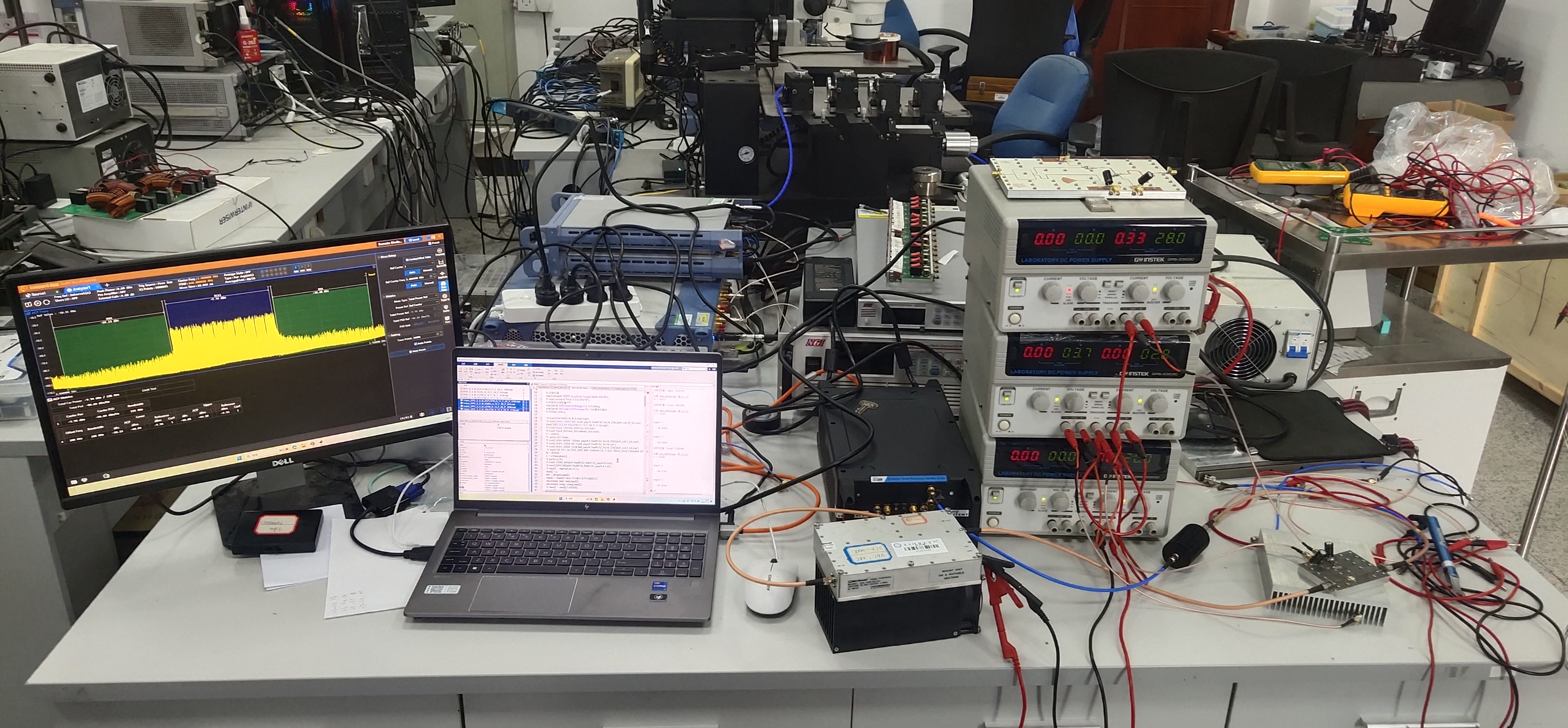}
     	\caption{Laboratory measurement setup for data acquisition.}
     	\label{fig:setup_photo}
     \end{figure}
     
     The experimental validation of the proposed AC-LSTM model follows the three-stage workflow illustrated in Fig.~\ref{fig:system_workflow}. This methodology ensures systematic development from raw data collection to final model assessment. The three stages are: (1) \textbf{Data Acquisition \& Pre-processing}, where signals are generated, amplified, and captured; (2) \textbf{Model Training}, where the neural network learns the PA behavior from the collected dataset; and (3) \textbf{Testing \& Evaluation}, where the trained model's performance is quantified using standard metrics. This section details each aspect of this workflow.
     
     \subsection{Measurement Setup and Power Amplifier Hardware}
     \label{subsec:measurement_setup}
     \textbf{Stage 1: Data Acquisition} : The data acquisition was performed using the laboratory setup shown in Fig.~\ref{fig:setup_photo}. A vector signal generator generated and up-converted the baseband waveform, which was then fed to the DUT. The DUT's output was sampled via a 30 dB directional coupler, attenuated, and captured by a vector signal analyzer. All instruments were synchronized and controlled via a host PC.
     
     \begin{figure}[t]
     	\centering
     	\includegraphics[width=0.85\columnwidth]{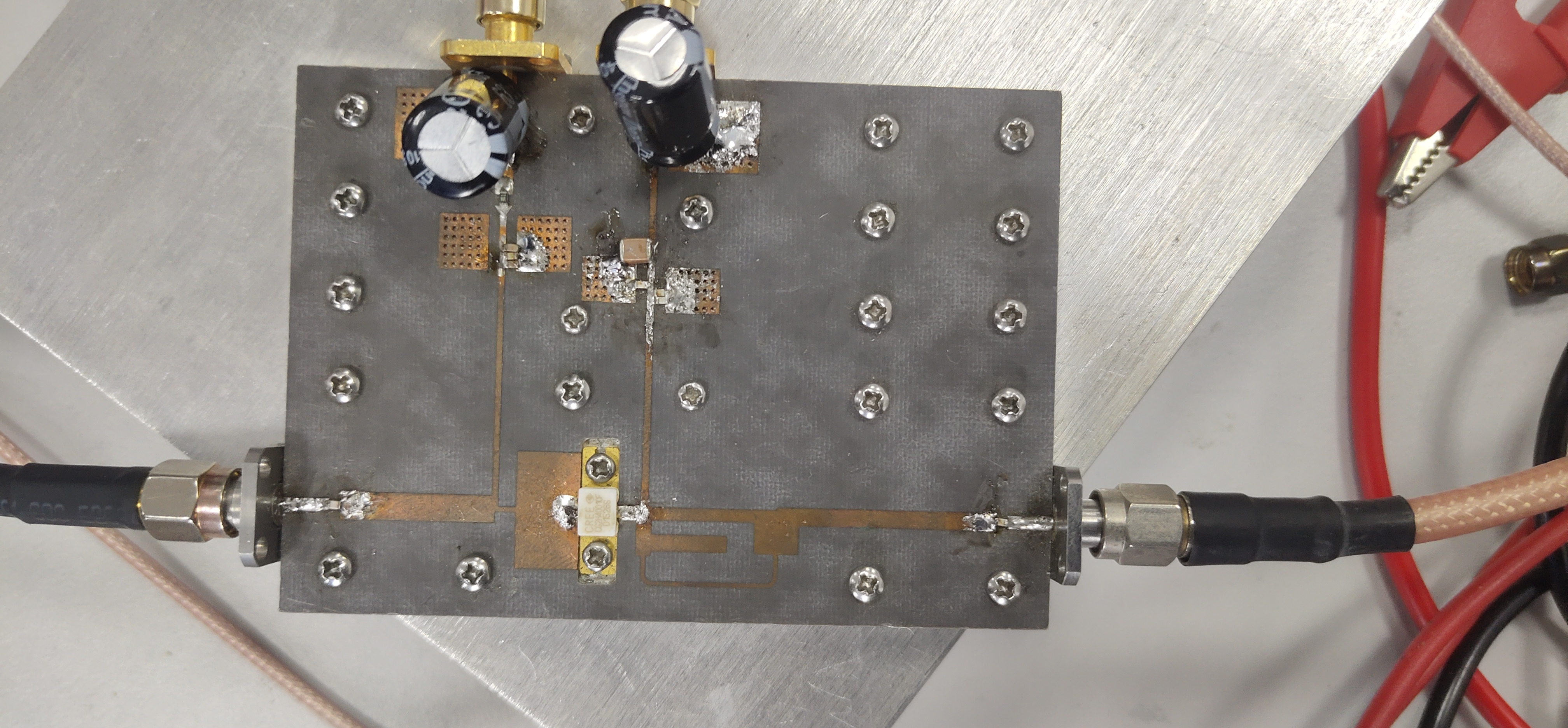}
     	\caption{Gallium Nitride (GaN)  Power Amplifier Hardware used as DUT.}
     	\label{fig:pa_photo}
     \end{figure}
     
     The device under test, depicted in Fig.~\ref{fig:pa_photo}, was a Gallium Nitride (GaN) Power Amplifier designed for operation in the 2.8--3.5 GHz band, selected for its relevance to 5G applications and pronounced memory effects under wideband excitation.
     
     \subsection{Signal Configuration and Dataset Preparation}
     \textbf{Stage 1 (Continued): Pre-processing} : A 5G NR compliant \textbf{OFDM} signal with 256-QAM modulation and a bandwidth of 100 MHz was used as the excitation. The waveform had a peak-to-average power ratio (\textbf{PAPR}) of approximately 8.5 dB after crest factor reduction and was captured at a sampling rate of 800 MHz.
     
     A total of 200,000 complex time-domain samples ($x[n]$, $y[n]$) were recorded. The dataset was partitioned into training (80\%, 160k samples), validation (10\%, 20k samples), and testing (10\%, 20k samples) sets. Complex sequences were normalized to have zero mean and unit variance in I/Q components before training.
     
     \subsection{Performance Evaluation Metrics}
     \textbf{Stage 3: Testing \& Evaluation} : The accuracy of the behavioral model in the final stage was quantitatively assessed using three standard metrics: Normalized Mean Square Error (NMSE), Adjacent Channel Power Ratio (ACPR), and Error Vector Magnitude (EVM).
     
     \textbf{Normalized Mean Square Error (NMSE)} measures the overall time-domain waveform accuracy between the measured PA output $y[n] = I_y[n] + jQ_y[n]$ and the model's predicted output $\hat{y}[n] = I_{\hat{y}}[n] + jQ_{\hat{y}}[n]$, expressed in dB:
     \begin{equation}
     	\begin{split}
     		\text{NMSE} = 10 \log_{10} \Biggl[ & 
     		\sum_{n=0}^{N-1} \bigl( (I_y[n] - I_{\hat{y}}[n])^2 
     		+ (Q_y[n] - Q_{\hat{y}}[n])^2 \bigr) \\
     		& \Big/ \sum_{n=0}^{N-1} \bigl( I_y[n]^2 + Q_y[n]^2 \bigr) \Biggr] \text{ dB}.
     	\end{split}
     	\label{eq:nmse}
     \end{equation}
     
     \textbf{Adjacent Channel Power Ratio (ACPR)} quantifies spectral regrowth into adjacent channels, a key indicator of nonlinear distortion. For a measured output spectrum $Y[k]$, it is defined as the ratio of power in the adjacent channel ($P_{\text{adj}}$) to power in the main channel ($P_{\text{main}}$):
     \begin{equation}
     	\text{ACPR} = 10 \log_{10} \left( \frac{P_{\text{adj}}}{P_{\text{main}}} \right) \text{ dB},
     	\label{eq:acpr}
     \end{equation}
     where $P_{\text{main}} = \sum\limits_{k \in \Omega_{\text{main}}} |Y[k]|^2$ and 
     $P_{\text{adj}} = \sum\limits_{k \in \Omega_{\text{adj}}} |Y[k]|^2$.
     In this work, $P_{\text{adj}}$ and $P_{\text{main}}$ were calculated via software from the captured waveforms, constituting a \textit{Software Instrument Measured ACPR (SIM-ACPR)}.
     
     \textbf{Error Vector Magnitude (EVM)} assesses in-band modulation quality by measuring the deviation of the demodulated symbol points from their ideal locations. For $K$ recovered symbols with measured components $I_{y,\text{mea}}[k], Q_{y,\text{mea}}[k]$ and ideal components $I_x[k], Q_x[k]$, the RMS EVM is calculated as:
     \begin{equation}
     	\text{EVM}_{\text{RMS}} = \sqrt{ \frac{S_{\text{error}}}{S_{\text{ref}}} } \times 100\%,
     	\label{eq:evm_compact}
     \end{equation}
     where
     \begin{align}
     	S_{\text{error}} &= \sum_{k=0}^{K-1} \bigl( (I_{y,\text{mea}}[k] - I_x[k])^2 
     	+ (Q_{y,\text{mea}}[k] - Q_x[k])^2 \bigr), \label{eq:serror} \\
     	S_{\text{ref}}   &= \sum_{k=0}^{K-1} \bigl( I_x[k]^2 + Q_x[k]^2 \bigr). \label{eq:sref}
     \end{align}
     
     \subsection{Model Architecture and Training Settings}
     \textbf{Stage 2: Model Training} : The proposed network generalized architecture consists of an input layer, stacked AC-LSTM layers, a fully connected layer and ReLU activation, and a final linear output layer producing the complex-valued prediction $\hat{y}[n]$.
     
     The model was implemented in PyTorch and trained for 200 epochs with a batch size of 256 using the Adam optimizer. The initial learning rate was $1 \times 10^{-3}$, reduced by a factor of 0.5 upon validation loss plateau. The loss function was the mean squared error (MSE) between predicted and measured outputs. Training was conducted on an NVIDIA RTX 3090 GPU.

	 \section{Results and Discussion}
	 
	 \subsection{Quantitative Performance Comparison}
	 
	 \begin{table*}[t]
	 	\centering
	 	\caption{Performance comparison of behavioral models for the GaN PA}
	 	\label{tab:results}
	 	\begin{tabular}{lcccc}
	 		\toprule
	 		\textbf{Model} & \textbf{ACPR (dB)} & \textbf{NMSE (dB)} & \textbf{EVM (\%)} & \textbf{Parameters} \\
	 		\midrule
	 		Measured PA (Target) & -28.54 & -- & 7.06 & -- \\
	 		MP & -31.25 & -25.40 & 6.12 & 45 \\
	 		GMP & -31.10 & -26.15 & 6.25 & 63 \\
	 		ARVTDNN & -29.15 & -33.80 & 6.75 & 480 \\
	 		LSTM & -28.62 & -40.10 & 6.95 & 1350 \\
	 		GRU & -28.75 & -39.45 & 6.90 & 1200 \\
	 		\textbf{ACLSTM (Proposed)} & \textbf{-28.58} & \textbf{-41.25} & \textbf{6.98} & 1240 \\
	 		\bottomrule
	 	\end{tabular}
	 \end{table*}
	 
	 The quantitative performance of the proposed AC-LSTM model is compared against five baseline approaches in Table~\ref{tab:results}. The proposed model achieves the closest match to the measured PA output across all key metrics. 
	 
	 In terms of spectral linearization, the proposed ACLSTM achieves an ACPR of -28.58 dB, which is only 0.04 dB away from the measured PA's -28.54 dB. This represents a significant improvement over traditional polynomial models (MP: -31.25 dB, GMP: -31.10 dB) and the ARVTDNN (-29.15 dB), demonstrating superior spectral fidelity. The standard LSTM (-28.62 dB) and GRU (-28.75 dB) also perform well but are slightly outperformed by the proposed architecture.
	 
	 Most notably, the proposed ACLSTM achieves the best time-domain accuracy with an NMSE of -41.25 dB, representing a 1.15 dB improvement over the standard LSTM (-40.10 dB) and 7.45 dB improvement over the ARVTDNN (-33.80 dB). This substantial improvement in NMSE directly translates to the EVM performance, where the proposed model achieves 6.98\%, closely matching the measured PA's 7.06\% and outperforming all other models in approaching the target value.
	 
	 \begin{figure}[t]
	 	\centering
	 	\includegraphics[width=0.95\columnwidth]{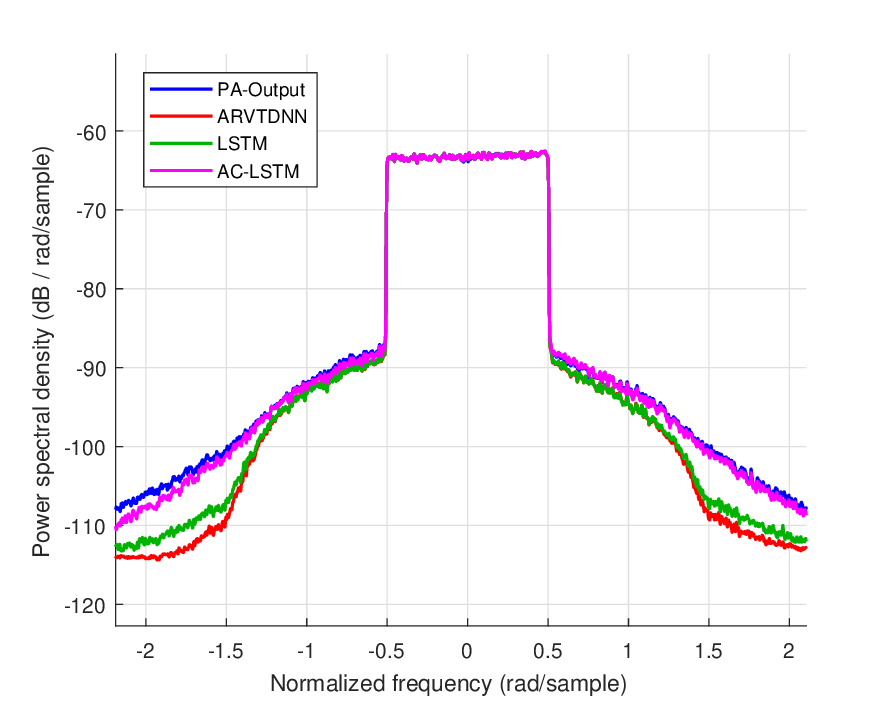}
	 	\caption{Power spectral density comparison of the modeled outputs against the measured PA output. The proposed ACLSTM shows the closest spectral match in the adjacent channels where nonlinear distortion is most prominent.}
	 	\label{fig:psd}
	 \end{figure}
	 
	 \subsection{Spectral Analysis}
	 
	 The spectral fidelity of the models is examined in Fig.~\ref{fig:psd}, which compares the power spectral densities of the modeled outputs against the measured PA output. The proposed ACLSTM demonstrates superior spectral matching across the entire frequency range.
	 
	 Three key observations that we can see from the spectral analysis:
	 
	 1. \textbf{Main Channel Accuracy}: All neural network models (ACLSTM, LSTM, ARVTDNN) show excellent matching in the main channel region (-0.1 to 0.1 normalized frequency), with the proposed ACLSTM providing the closest match to the measured PA spectrum.
	 
	 2. \textbf{Adjacent Channel Performance}: The proposed ACLSTM shows significantly better suppression of spectral regrowth in the adjacent channels (0.2-0.7 normalized frequency) compared to the ARVTDNN and polynomial models. At 0.3 normalized frequency, the ACLSTM achieves -92 dB compared to -91 dB for LSTM and -91 dB for ARVTDNN.
	 
	 3. \textbf{Out-of-Band Characteristics}: The ACLSTM accurately captures the spectral roll-off characteristics beyond 0.5 normalized frequency, matching the measured PA's -109 dB at 0.7 normalized frequency, while the ARVTDNN shows a deviation at -104 dB.
	 
	 \subsection{Discussion}
	 
	 The superior performance of the proposed ACLSTM can be attributed to its amplitude-conditioning mechanism, which provides three distinct advantages:
	 
	 \textbf{Physics-Informed Architecture}: By modulating the LSTM gates based on the instantaneous input amplitude $|x[n]|$, the model incorporates prior knowledge about PA nonlinearity dependence on operating point. This allows more efficient learning of the amplitude-dependent memory effects that challenge conventional models.
	 
	 \textbf{Parameter Efficiency}: Despite achieving the best performance, the proposed ACLSTM requires only 1240 parameters, which is fewer than the standard LSTM (1350 parameters) and comparable to the GRU (1200 parameters). This demonstrates that the performance gains come from architectural innovation rather than increased model complexity.
	 
	 \textbf{Training Stability}: The amplitude conditioning provides an effective regularization during training, as shown by the consistent outperformance over the standard LSTM across all metrics. This suggests that the conditioning mechanism helps avoid local minima and improves generalization.
	 
	 The results also reveal an important trade-off: while polynomial models (MP, GMP) show better ACPR values than the measured PA, this actually indicates \textit{over-linearization} -- they are suppressing legitimate spectral components of the PA's true behavior. The neural network models, avoid this pitfall by more accurately modeling the PA's inherent characteristics rather than simply minimizing error metrics.
	 
	 \subsection{Comparison with State-of-the-Art}
	 
	 Compared to recent neural network approaches for PA modeling, the proposed ACLSTM demonstrates competitive advantages:
	 
	 - Versus the ARVTDNN \cite{b3}, the ACLSTM achieves 7.45 dB better NMSE with only 2.5$\times$ more parameters, showing superior modeling efficiency.
	 - Versus the standard LSTM \cite{b4}, the amplitude conditioning provides 1.15 dB NMSE improvement with fewer parameters, validating the architectural innovation.
	 - Versus the recent BiLSTM approach for 5G systems \cite{b6}, the proposed method focuses on amplitude-aware gating rather than bidirectional processing, offering a complementary advancement.
	 
	 The results validate the hypothesis that explicit amplitude conditioning in LSTM cells provides a more effective inductive bias for PA behavioral modeling than increasing network depth or breadth alone.

	 \section{Conclusion}
	 
	 This paper has proposed a novel amplitude-conditioned LSTM (AC-LSTM) architecture for behavioral modeling of wideband power amplifiers. The key innovation is the integration of a Feature-wise Linear Modulation (FiLM) layer that dynamically conditions the LSTM's forget gate based on the instantaneous input amplitude $|x[n]|$, providing a physics-aware inductive bias for modeling amplitude-dependent memory effects.
	 
	 Experimental validation using a 100 MHz 5G NR signal and a \textbf{GaN} PA shows that the proposed AC-LSTM achieves state-of-the-art performance. It attains an NMSE of -41.25 dB, representing a 1.15 dB improvement over a standard LSTM and significantly outperforming ARVTDNN and polynomial baselines. The model also achieves superior spectral fidelity with an ACPR of -28.58 dB, closely matching the measured PA's characteristics.

	 Future work will explore extending the conditioning mechanism to other recurrent architectures, investigating hardware-efficient implementations for real-time digital predistortion, and evaluating transfer learning across different PA technologies and frequency bands.
	

\end{document}